\documentclass{article}

\usepackage{PRIMEarxiv}

\usepackage[utf8]{inputenc} 
\usepackage[T1]{fontenc}    
\usepackage{hyperref}       
\usepackage{url}            
\usepackage{booktabs}       
\usepackage{amsfonts}       
\usepackage{nicefrac}       
\usepackage{microtype}      
\usepackage{lipsum}
\usepackage{fancyhdr}       
\usepackage{graphicx}       
\graphicspath{{media/}}     

\title{
Multivariate Beta Mixture Model: Probabilistic Clustering With Flexible Cluster Shapes
}

\author{
  Yung-Peng Hsu, Hung-Hsuan Chen \\
  National Central University \\
  Taoyuan, Taiwan\\
  \texttt{yungpeng1998@gmail.com, hhchen1105@acm.org} \\
}

\usepackage{algorithm2e}
\usepackage{amsfonts}
\usepackage{amsmath}
\usepackage{booktabs}
\usepackage{color}
\usepackage{graphicx}
\usepackage{makecell}
\usepackage{tabularx}

\newcommand{\figwidth}[4]{
    \begin{figure}[tb]\centering
    \includegraphics[width=#4]{./fig/#1}
    \caption{#2}
    \label{#3}\end{figure}
    {}}
    
\newcommand{\norm}[1]{
    \left\lVert#1\right\rVert}

\newcommand{\figccwidth}[4]{
    \begin{figure*}[tb]\centering
    \includegraphics[width=#4]{./fig/#1}
    \caption{#2}
    \label{#3}\end{figure*}
    {}}

\begin{document}
\maketitle

\begin{abstract}

This paper introduces the multivariate beta mixture model (MBMM), a new probabilistic model for soft clustering. MBMM adapts to diverse cluster shapes because of the flexible probability density function of the multivariate beta distribution. We introduce the properties of MBMM, describe the parameter learning procedure, and present the experimental results, showing that MBMM fits diverse cluster shapes on synthetic and real datasets.  The code is released anonymously at \url{https://github.com/hhchen1105/mbmm/}.

\keywords{Mixture model  \and EM algorithm \and Clustering.}
\end{abstract}

\section{Introduction}

Data clustering groups data points into components so that similar points are within the same component. Data clustering is commonly used for data exploration and is sometimes used as a preprocessing step for later analysis~\cite{lien2019visited}.  

In this paper, the multivariate beta mixture model (MBMM), a new probabilistic model for soft clustering, is proposed. As the MBMM is a mixture model, it shares many properties with the Gaussian mixture model (GMM), including its soft cluster assignment and parametric modeling. In addition, the MBMM allows the generation of new (synthetic) instances based on a generative process. Because the beta distribution is highly flexible (e.g., unimodal, bimodal, straight line, or exponentially increasing or decreasing), MBMM can fit data with versatile shapes. Figure~\ref{fig:bivariate-beta-exp} shows that various cluster shapes can be obtained with a bivariate beta distribution. On the contrary, the shape of a Gaussian distribution is symmetric and unimodal, which limits its fitting capacity.

\figwidth{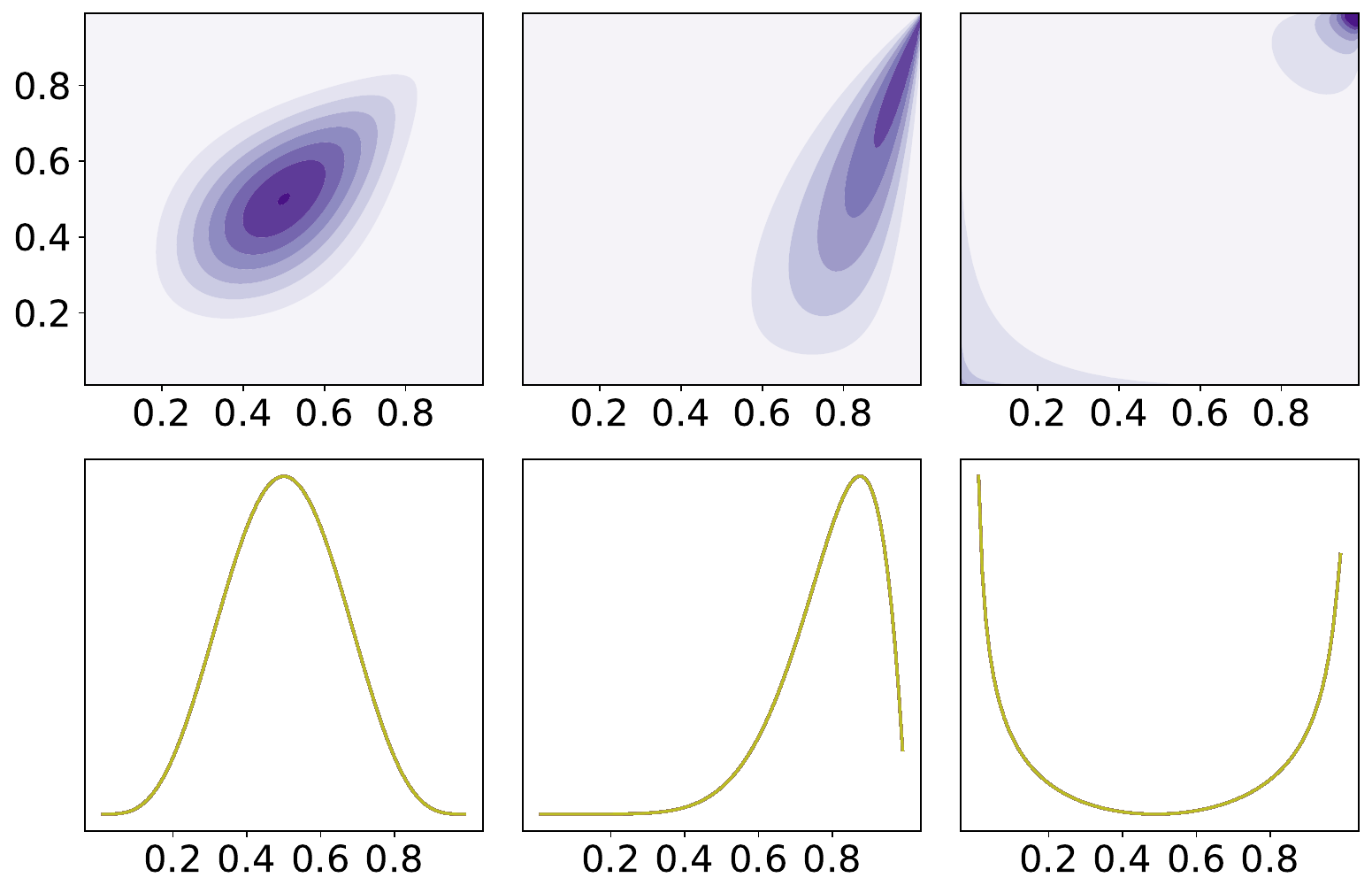}
{Examples of the versatile shape of the bivariate beta distribution. The upper row shows three bivariate beta distributions with different parameters. The bottom row shows the marginal distribution of $x_1$ (i.e., the variable on the horizontal axis in the top row). This distribution can be symmetric unimodal (e.g., the left subfigure), skewed unimodal (e.g., the middle subfigure), or bimodal (e.g., the right subfigure).}
{fig:bivariate-beta-exp}{.6\columnwidth}

The multivariate beta distribution is defined in different ways. In some studies, the Dirichlet distribution is considered a multivariate beta distribution (e.g.,~\cite{kotz2004continuous}) because the beta distribution is a special case of the Dirichlet distribution with two parameters. However, we apply the definition provided in~\cite{jones2002multivariate}, which is even more general than the Dirichlet distribution. The relationship between the Dirichlet distribution and our multivariate beta distribution will be discussed in Section~\ref{sec:multi-beta} when we introduce the details of the multivariate beta distribution.

This paper presents several contributions. First, we propose a new probabilistic model for soft clustering. Our model is similar to the Gaussian Mixture Model (GMM), but the shape of each cluster is more versatile than those generated by GMM. Second, we compare MBMM with well-known clustering algorithms on synthetic and real datasets to demonstrate its effectiveness. Finally, we release the code for reproducibility. Our implemented class offers \texttt{fit()}, \texttt{predict()}, and \texttt{predict\_proba()}, the common methods provided by \texttt{scikit-learn}'s clustering algorithms, making it convenient to apply MBMM to new domains.

The rest of the paper is organized as follows. Section~\ref{sec:method} introduces the multivariate beta distribution and the proposed MBMM. Section~\ref{sec:exp} describes experiments on synthetic and real datasets. Section~\ref{sec:rel-work} reviews previous work on data clustering. We conclude by discussing the limitations of the MBMM and the ongoing and future work on the MBMM in Section~\ref{sec:disc}.

\section{Multivariate Beta Mixture Model} \label{sec:method}


\subsection{Multivariate beta distribution} \label{sec:multi-beta}

The probability density function (PDF) of a multivariate beta distribution (MB) has been defined in different ways~\cite{arnold2011flexible,jones2002multivariate}. Here, we apply the definition in~\cite{jones2002multivariate}: given an instance $\boldsymbol{x} = [x_1, \ldots, x_M]^T$ with $M$ variates (i.e., features) and the shape parameters $a_m > 0, b > 0~(m=1, \ldots, M)$, its PDF is given by
\begin{equation} \label{eq:beta-dist}
    \mathit{MB}(\boldsymbol{x}|a_{1:M}, b) = \frac{1}{Z} \times \frac{\prod_{m=1}^M \frac{x_m^{a_m-1}}{(1-x_m)^{a_m+1}}}{(1+\sum_{k=1}^M\frac{x_k}{1-x_k})^{a_1+\ldots+a_M+b}},
\end{equation}
where $x_m \in (0,1), a_m > 0, b > 0$, and the normalizer~$Z$ is defined by
\begin{equation} \label{eq:beta-dist-normalizer}
    Z = \frac{\Gamma(b)\prod_{m=1}^M\Gamma(a_m)}{\Gamma(b + \sum_{j=1}^M a_j)},
\end{equation}
where $\Gamma$ is the gamma function.

\figwidth{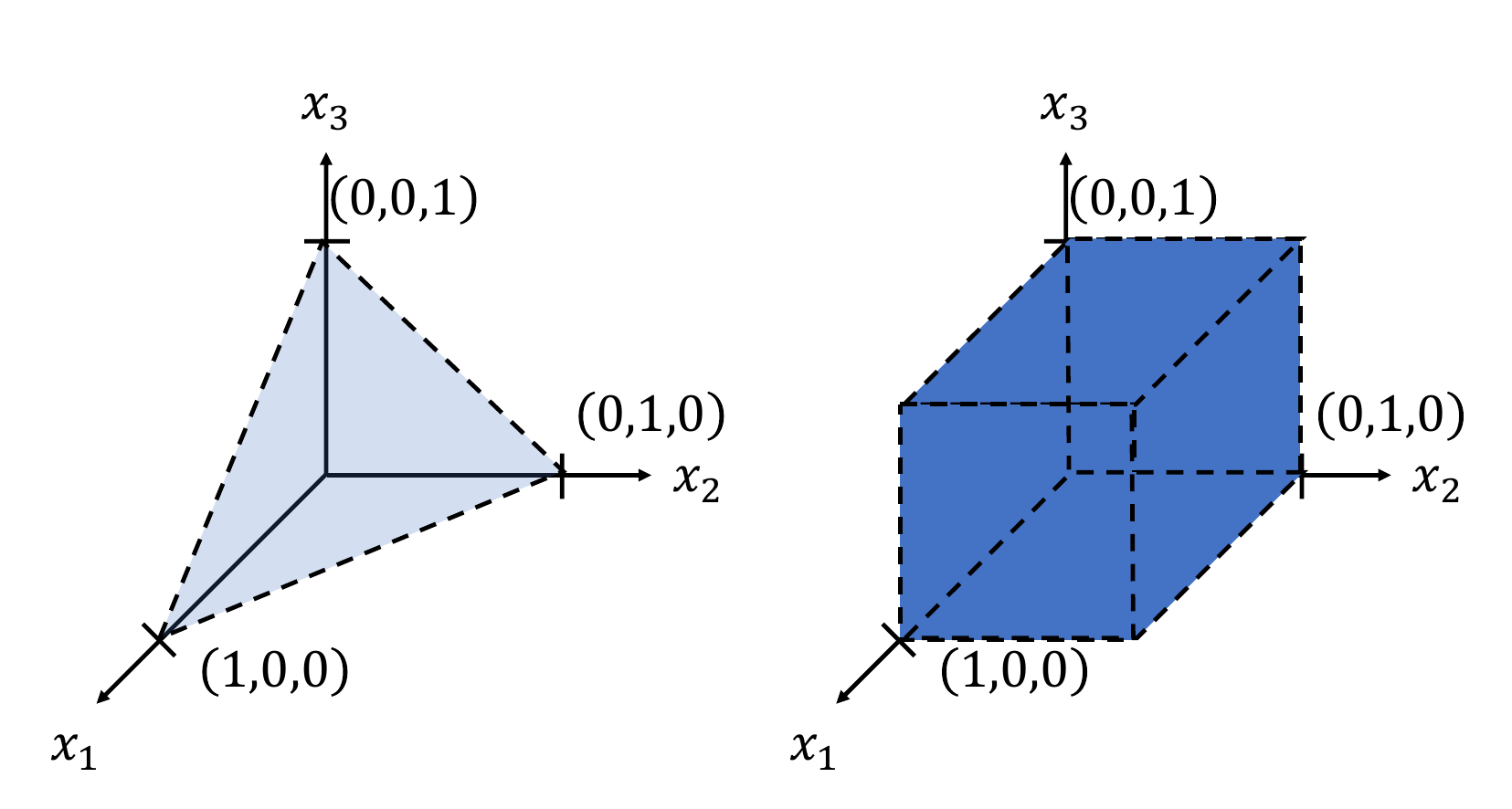}{A comparison of the support of the Dirichlet distribution (left) and our multivariate beta distribution (right) with 3 variates.  The Dirichlet distribution is only defined on $x_i \in (0, 1)$ such that $x_1 + x_2 + x_3 = 1$ (the standard 2-simplex in $R^3$).  On the contrary, our multivariate beta distribution is defined on $(0,1)^3$ (the unit cube in $R^3$), which is a superset of the Dirichlet distribution.}{fig:dirichlet-vs-multiv-beta}{.6\columnwidth}

In some previous studies, the Dirichlet distribution was treated as a multivariate generalization of the beta distribution (e.g.,~\cite{kotz2004continuous}) since the Dirichlet distribution falls back to the beta distribution when the number of parameters is 2. However, we describe a more general definition of the multivariate beta distribution that regards the Dirichlet distribution as a special case. The relationship between the Dirichlet distribution and the proposed multivariate beta distribution is illustrated in Figure~\ref{fig:dirichlet-vs-multiv-beta}. Specifically, the support of an $n$-variate Dirichlet distribution is restricted to a standard $(n-1)$-simplex. However, the support of our multivariate beta distribution is a hypercube in an $n$-dimensional space with a length of 1 on each side. In other words, the Dirichlet distribution is the multivariate beta distribution subject to $\norm{\boldsymbol{x}}_1 = 1$.

\subsection{MBMM density function and generative process}

In Table~\ref{tab:var-list}, we list the notations that will be used in this paper hereafter.

\begin{table}[tb]
\caption{Notation list}
\label{tab:var-list}
\begin{center}
\begin{tabularx}{\columnwidth}{p{0.05\columnwidth}p{0.08\columnwidth}X}
\toprule
  
\multicolumn{3}{l}{{Indices:}} \\
  & $M$ & Dimensions of an observed instance ($m \in \{1, \ldots, M\}$) \\
  & $C$ & Number of clusters ($c \in\{1, \ldots, C\}$) \\ 
  & $N$ & Number of instances ($n \in\{1, \ldots, N\}$) \\
  
\multicolumn{3}{l}{{Parameters:}}      \\  
  & $a_{c,m}$ & the $m$-th shape parameter for cluster $c$; $a_{c,m} > 0$ \\
  & $b_{c}$ & the $(M+1)$-th shape parameter for cluster $c$; $b_c > 0$ \\
  & $\pi_{c}$ & Mixture weight of cluster $c$, $0 < \pi_c < 1, \sum_{c=1}^C \pi_c = 1$ \\
  & $z_n$ & Cluster that $\boldsymbol{x}_n$ belongs to; $z_n \in \{1, \ldots, C\}$ \\
  & $\gamma_{n,c}$ & Probability that $\boldsymbol{x}_n$ belongs to cluster $c$; $\sum_{c=1}^C \gamma_{n,c} = 1$ \\
  & $\boldsymbol{\theta}_c$ & Set of parameters for cluster $c$;  $\boldsymbol{\theta}_c = \left\{a_{c,1}, \ldots, a_{c,M}, b_c \right\}$ \\
  & $\boldsymbol{\theta}$ & Set of parameters for the MBMM;  $\boldsymbol{\theta} = \left\{a_{1:C,1:M}, b_{1:C}, \pi_{1:C}\right\}$ \\
\multicolumn{3}{l}{{Observed random variables:}}   \\  
  & $\boldsymbol{x}_n$ & Observed instance; $\boldsymbol{x}_n = \left[x_{n,1}, \ldots, x_{n,M}\right]^T \in R^M$ \\
  \bottomrule
 \end{tabularx}
\end{center}
\end{table}

In MBMM, it is assumed that the data points are generated from a mixture of multivariate beta distributions (whose PDF is defined in Equation~\ref{eq:beta-dist}). Consequently, the probability of the MBMM given $C$ components is 
\begin{equation} \label{eq:mbmm}
p(\boldsymbol{x}_n | \boldsymbol{\theta}) = \sum_{c=1}^C \pi_c \mathit{MB}(\boldsymbol{x}_n|\boldsymbol{\theta}_c).
\end{equation}

The parameter $\pi_c$ determines the probability that a random instance $\boldsymbol{x}_n$ belongs to cluster $c$ (before knowing the values of the variates in $\boldsymbol{x}_n$), and $\mathit{MB}(\boldsymbol{x}_n|\boldsymbol{\theta}_c)$ gives the PDF if $\boldsymbol{x}_n$ indeed belongs to cluster $c$.

\figwidth{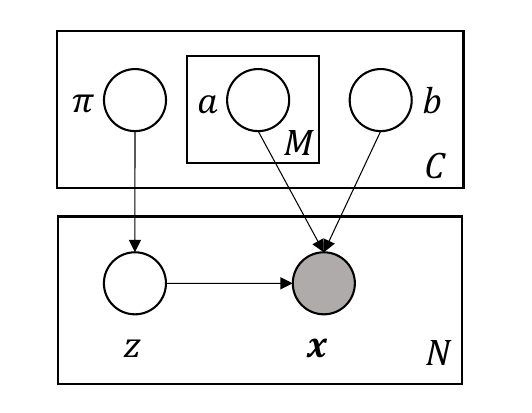}{Graphical representation of the multivariate
beta mixture model}{fig:mbmm-plate-notation}{.32\columnwidth}

Figure~\ref{fig:mbmm-plate-notation} shows a graphical representation of the multivariate beta mixture model. To generate a sample $\boldsymbol{x}_n$, we first sample a latent variable $z_n$ (the cluster ID of the sample $\boldsymbol{x}_n$) from a multinomial distribution with parameters $\pi_1, \ldots, \pi_C$. Suppose that $z_n=c$ after sampling, we further sample an instance $\boldsymbol{x}_n$ from the multivariate beta distribution with parameters $\boldsymbol{\theta}_c$: $\mathit{MB}(\boldsymbol{x}_n|\boldsymbol{\theta}_c) = \mathit{MB}(\boldsymbol{x}_n|a_{c,1}, \ldots, a_{c,M}, b_c)$.

\subsection{Parameter Learning for the MBMM}

\begin{algorithm}
\caption{Parameter learning algorithm for the MBMM}\label{alg:mbmm-learn}

\KwData{Input data $\boldsymbol{x}_1, \ldots, \boldsymbol{x}_N$, cluster number $C$}

\KwResult{Parameters $\boldsymbol{\theta}=\{\pi_{1:C}, a_{1:C, 1:M}, b_{1:C}\}$}

Initialize $\boldsymbol{\theta}$ randomly\;

\While{\textup{not converge}} {
    \tcp{E-step}
    \For{$n \leftarrow 1$ \KwTo N}{
        \For{$c \leftarrow 1$ \KwTo C}{
            Update $\gamma_{n,c}$ by Equation~\ref{eq:gamma-nc}\;
        }
    }
    
    \tcp{M-step}
    Update $a_{1:C, 1:M}$ and $b_{1:C}$ with the SQP solver~\cite{kraft1988software}\;
    Update $\pi_{1:C}$ by Equation~\ref{eq:pi_j}\;
    
    \If{\textup{iteration count reaches a pre-defined value}}{
        Exit while loop\;
    }
}
\end{algorithm}

In reality, we do not know the values of the parameters $\boldsymbol{\theta} = \left\{a_{1:C,1:M}, b_{1:C}, \pi_{1:N}\right\}$ (referring to Figure~\ref{fig:mbmm-plate-notation}). We hope to recover these parameters based on the observed $\boldsymbol{x}_n$-s to maximize the likelihood function:
\begin{equation} \label{eq:mbmm-n-instances}
L(\boldsymbol{\theta}) = p(\boldsymbol{x}_{1:N}, z_{1:N} | \boldsymbol{\theta}) = \prod_{n=1}^N\prod_{c=1}^C \left[\pi_c \mathit{MB}(\boldsymbol{x}_n | \boldsymbol{\theta}_c)\right]^{I(z_n=c)},
\end{equation}
where $I$ is the indicator function.

As the likelihood function (Equation~\ref{eq:mbmm-n-instances}) involves the multiplication of $N \times C$ terms, the result is numerically unstable. Instead, we compute the log-likelihood function to convert multiplications to additions, as shown in Equation~\ref{eq:loglikelihood-mbmm}. As a result, the computation is more numerically stable.
\begin{equation} \label{eq:loglikelihood-mbmm}
\log L(\boldsymbol{\theta}) = \sum_{n=1}^N \sum_{c=1}^C I(z_n=c) \left(\log \pi_c + \log MB(\boldsymbol{x}_n | \boldsymbol{\theta}_c)\right).
\end{equation}

However, since we cannot observe the latent $z_n$ in practice, direct optimization of Equation~\ref{eq:loglikelihood-mbmm} is difficult. As an alternative, we compute the expected value of the log-likelihood function with respect to the latent variables $z_{1:N}$, which involves the expected (but not the true) values of $z_n$:
\begin{equation} \label{eq:exp-loglikelihood-mbmm}
E_{z_{1:N}}\left[\log L(\boldsymbol{\theta})\right] = \sum_{n=1}^N \sum_{c=1}^C \gamma_{n,c} \left(\log \pi_c + \log MB(\boldsymbol{x}_n | \boldsymbol{\theta}_c)\right).
\end{equation}

After the above reformulation, the parameters ($\boldsymbol{\theta}_{1:C}, \pi_{1:C}$) that are used to maximize the expected value of the log-likelihood function (Equation~\ref{eq:exp-loglikelihood-mbmm}) can be learned via the EM algorithm, as given by the pseudocode in Algorithm~\ref{alg:mbmm-learn}. In the E-step, we compute $\gamma_{n,c}$ (the probability that instance $\boldsymbol{x}_n$ belongs to cluster~$c$) that maximizes Equation~\ref{eq:exp-loglikelihood-mbmm} by assuming that the randomly initialized or currently estimated $\pi_{1:C}$ and $\boldsymbol{\theta}_{1:C}$ are correct.  The assignment of $\gamma_{n,c}$ has a simple closed-form solution, as shown below
\begin{equation} \label{eq:gamma-nc}
    \gamma_{n,c} = \frac{\pi_c \mathit{MB}(\boldsymbol{x}_n|\boldsymbol{\theta}_c)}{\sum_{k=1}^C \pi_k \mathit{MB}(\boldsymbol{x}_n|\boldsymbol{\theta}_k)}.
\end{equation}

In the M-step, we search for the parameters $\pi_{1:C}$, $a_{1:C,1:M}$, and $b_{1:C}$ by assuming that the estimated $\gamma_{n,c}$ values in the E-step are correct.  However, since the $a_{1:C,1:M}, b_{1:C}$ parameters seem to lack a closed-form solution, we resort to numerical optimization strategies, specifically the sequential quadratic programming (SQP) iterative method, as the minimization strategy~\cite{kraft1988software} because SQP allows linear constraints on the parameters (i.e., $a_{c,m} > 0 \textrm{ and } b_c > 0~\forall c, m$). For parameters $\pi_1, \ldots, \pi_C$, we rely the efficient closed-form solution:
\begin{equation} \label{eq:pi_j}
    \pi_c = \frac{1}{N}\sum_{n=1}^N \gamma_{n,c}.
\end{equation}

We compute the difference between the log-likelihood estimation in successive rounds for the convergence check. Additionally, if the number of iterations reaches a predefined value, we terminate the loop.

\subsection{The similarity score between data points} \label{sec:sim-func}

Most clustering algorithms define the distance between two samples by converting them into a non-negative real value, i.e., given $\boldsymbol{x}_i, \boldsymbol{x}_j \in R^M$, the distance function is represented by $d_{i,j}: \boldsymbol{x}_i \times \boldsymbol{x}_i \xrightarrow{} \{0, R^+\}$. However, if we define the distance of two samples based on their coordinates and assign samples to the closest cluster centroid, the output shapes of the clusters are inevitably convex.

In MBMM, we define the distance between two data points from a different perspective. Since the PDF of a data point is an affine combination of $C$ multivariate beta distributions (Equation~\ref{eq:mbmm}), we consider $MB(\cdot|\boldsymbol{\theta}_1), \ldots, MB(\cdot|\boldsymbol{\theta}_C)$ as the basis to form a function space. Consequently, the coordinate of the data point $\boldsymbol{x}_n$ becomes $\boldsymbol{\gamma}_n = \left[\gamma_{n,1}, \gamma_{n,2}, \ldots, \gamma_{n, C}\right]^T$ with respect to the basis functions. The vector $\boldsymbol{\gamma}_n$ is a discrete probability distribution since $\sum_{c=1}^C \gamma_{n,c} = 1$. Thus, we can define the distance between the data points $\boldsymbol{x}_i$ and $\boldsymbol{x}_j$ as the distance between the discrete probability distributions $\boldsymbol{\gamma}_i$ and $\boldsymbol{\gamma}_j$. We use the Kullback-Leibler divergence (KL divergence) to determine this distance:

\begin{equation} \label{eq:kl-diverg}
    d_{i,j}^{\mathit{KL}} := \sum_{c=1}^C \gamma_{i,c} \log \left( \frac{\gamma_{i,c}}{\gamma_{j,c}} \right).
\end{equation}

\section{Experiments} \label{sec:exp}

\subsection{Comparisons on the synthetic datasets}
\label{sec:syn-data-cluster}

\figccwidth{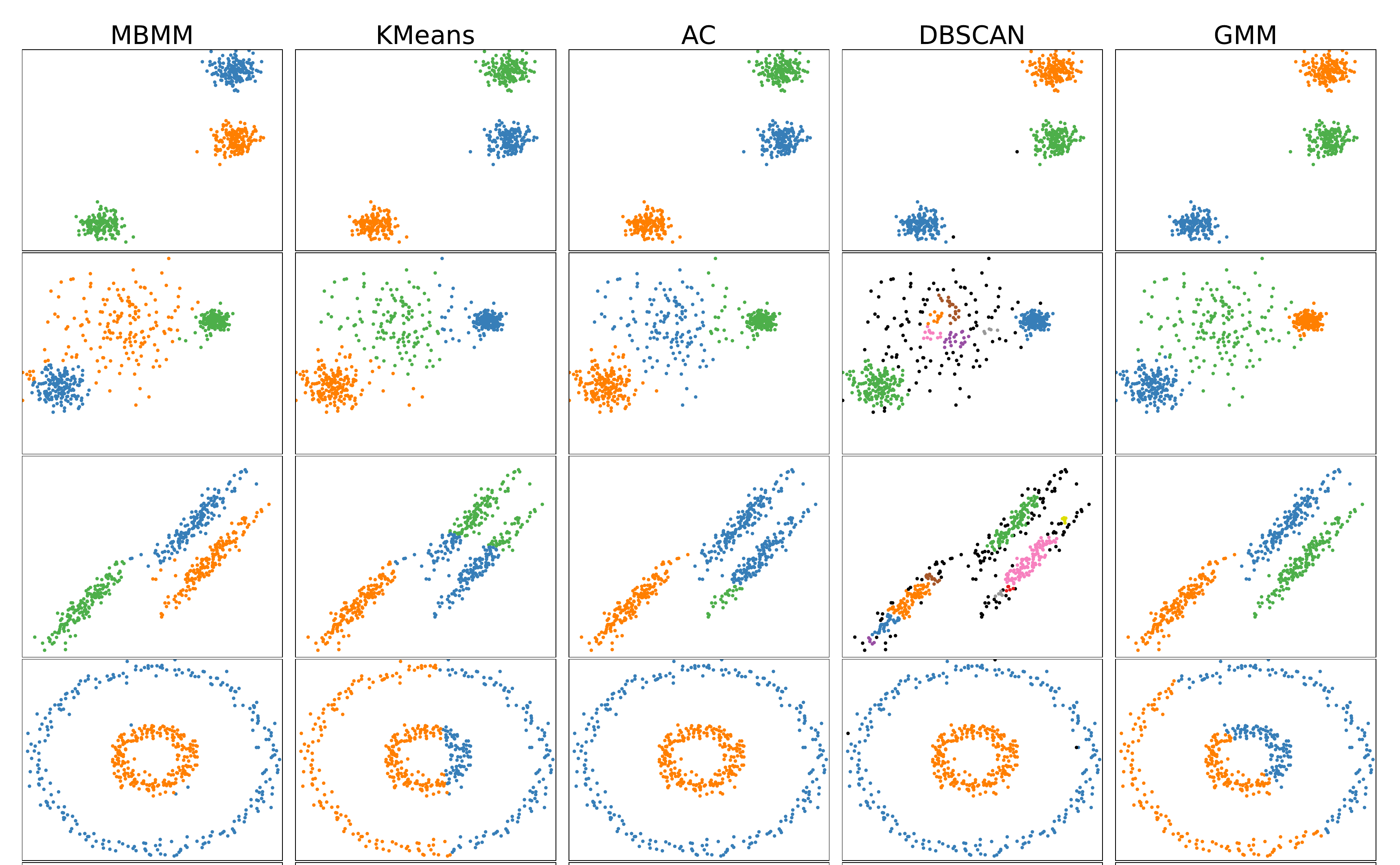}{Clustering algorithms on 
synthetic datasets}{fig:visualize-synthetic-cluster}{1.\linewidth}

\subsubsection{Baseline models}

Clustering algorithms can be classified into four types: centroid-based, density-based, distribution-based, and hierarchical clustering algorithms.  We select representative models from each of these categories.  For centroid-based models, we choose the $k$-means algorithm, which is likely the most widely used clustering algorithm.  We select the DBSCAN algorithm for density-based models, which received the Test of Time award at KDD 2014.  We choose the GMM as the distribution-based model and the agglomerative clustering (AC) algorithm as the hierarchical clustering model.

\subsubsection{Synthetic dataset generation}

We generate synthetic datasets for the experiments. First, we create data points from three isotropic 2D Gaussian distributions whose means are distant and whose variances are small in each dimension.  Consequently, a data point is close to other data points of the same Gaussian distribution but far from those in other Gaussian distributions (Figure~\ref{fig:visualize-synthetic-cluster}, first row).  This dataset represents an ideal case for data clustering.  

Second, we generate two distant 2D Gaussian distributions with small variances in each dimension. However, the third distribution has a large variance. Consequently, several data points sampled from the third distribution are mixed with those from other distributions (Figure~\ref{fig:visualize-synthetic-cluster}, second row).

Third, we generate three 2D Gaussian distributions with isolated means as before.  However, we introduce a high correlation between two covariates for each Gaussian distribution.  Consequently, data points are sometimes closer to data points generated from other distributions (Figure~\ref{fig:visualize-synthetic-cluster}, third row).

Fourth, we generate concentric circles (i.e., one circle within another).  If a point from the outer circle is selected, the most distant data point is located on the other side of the same circle. Consequently, such a synthetic dataset is highly challenging for centroid-based and distribution-based methods to group the outer circle as one cluster (Figure~\ref{fig:visualize-synthetic-cluster}, fourth row).

We scale the range of every feature to within 0 and 1 because the support of each variate in the MB distribution is between 0 and 1, as explained in Section~\ref{sec:multi-beta}.  This constraint has little influence in practice because clustering algorithms generally require the availability of all data sets prior to training; thus, scaling each variate as a preprocessing step is straightforward.

\subsubsection{Visualizing clustering results}

\figccwidth{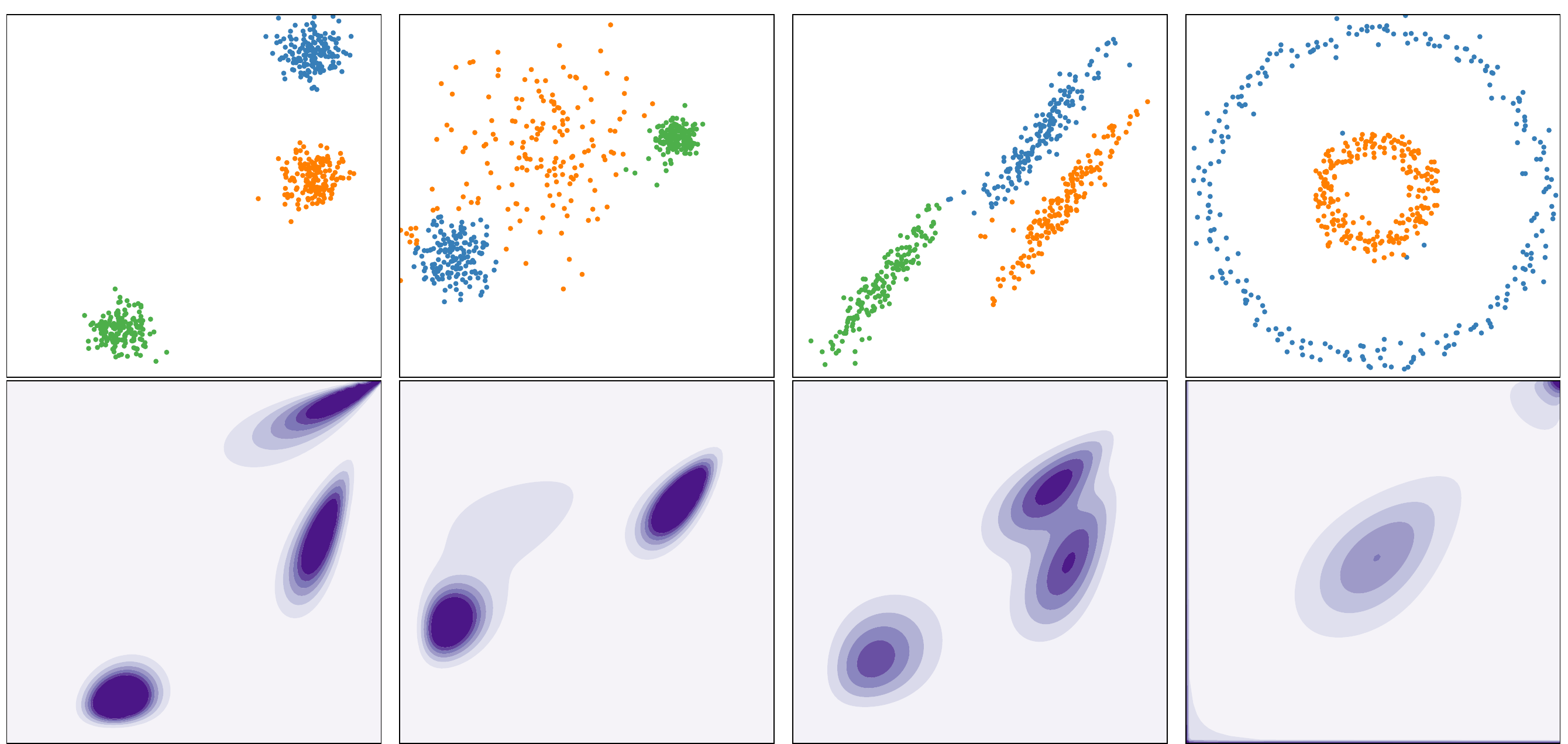}{Upper row: data points clustered by the MBMM; bottom
row: PDFs based on the fitted parameters.}{fig:mbmm-pdf}{.8\textwidth}

Figure~\ref{fig:visualize-synthetic-cluster} shows a visualized comparison of the clustering algorithms for the four synthetic datasets.  

All algorithms compared perform well on the first synthetic dataset.  However, some data points belonging to the middle cluster are incorrectly grouped as the right cluster for the second dataset when using $k$-mean and AC.  This is because the middle cluster has a wider spread, making the centroid far away from some points in the same cluster.  With DBSCAN, many data points from the middle cluster are regarded as outliers because density-based algorithms usually have difficulty when the intracluster distance (the distance between members of a cluster) differs greatly.  For similar reasons, unsatisfied performance is obtained using the $k$ means and DBSCAN on the third dataset.  Mediocre results are also obtained using the AC algorithm, likely because Ward's linkage function merges the wrong groups.  On the concentric circles dataset, poor performance is observed for the $k$-means and GMM algorithms because they can only group geometrically neighboring nodes in one cluster.  Since the AC and DBSCAN algorithms can recursively group adjacent points into the same cluster, it is possible to group two distant points into the same cluster, so reasonable performance is achieved. Excellent results are obtained using our proposed MBMM on all synthetic datasets because the shape of a multivariate beta distribution is versatile.  In particular, for the fourth dataset, because a multivariate beta distribution can be bimodal, the MBMM can group the data points in the outer circle as one cluster even though they are geometrically distant.

We visualize the PDFs by fitting the MBMM to the four synthetic datasets (Figure~\ref{fig:mbmm-pdf}). The upper row shows the data points, and the bottom row shows the PDFs estimated by MBMM, which indeed fits these datasets adequately.

\subsection{Comparison on the real datasets}

\subsubsection{Real datasets}

We used two open real datasets. The first dataset is MNIST, which includes grayscale images of handwritten digits. The size of each image is $28 \times 28$.  Since image pixels should have spatial correlations, directly using the pixel values as input features for clustering algorithms could be problematic.  Eventually, we reduce the dimension of each image to 2 dimensions using the following procedure.  First, we train a vanilla convolutional neural network (ConvNet) using the Fashion MNIST dataset (not the MNIST dataset). Then, we feed each MNIST image into the Fashion MNIST-trained ConvNet and take the hidden layer before the output (a vector with 512 neurons) as the image representation.  Finally, we use a standard autoencoder to reduce the vector into 2 dimensions, which are the inputs of the clustering algorithms.  Ultimately, we include only images of number 1 and number 9 in MNIST for experiments.

The second dataset, the breast cancer Wisconsin (diagnostic) dataset, consists of 569 instances.  Each instance includes 32 attributes and a binary class label indicating the status of the tumor (benign or malignant).  We download the dataset from the UCI Machine Learning Repository.


\subsubsection{Results}

We compare clustered IDs with ground truth labels to calculate the adjusted Rand index (ARI)~\cite{rand1971objective,wagner2007comparing} and adjusted mutual information (AMI)~\cite{vinh2010information}, two standard metrics for clustering evaluation.  If a clustering result perfectly matches the referenced clusters (labels), both metrics return a score of 1.  However, ARI and AMI are biased toward different types of clustering results: ARI prefers balanced partitions (clusters with similar sizes), and AMI prefers unbalanced partitions~\cite{romano2016adjusting}.  For a fair comparison, we report both metrics.

\begin{table}[tb]
\caption{Comparison of the clustering algorithms on the MNIST dataset and the breast cancer dataset (mean $\pm$ standard deviation)}
\label{tab:mnist-cancer-cmp}
\begin{center}

\begin{tabular}{c||cc||cc}
\toprule
 & \multicolumn{2}{c||}{MNIST} & \multicolumn{2}{c}{breast cancer} \\ 
 & ARI & AMI & ARI & AMI \\
\hline\hline
MBMM & $\boldsymbol{.937}\pm .000$ & $\boldsymbol{.884}\pm .000$ & $\underline{.664} \pm .000$ & $\underline{.558} \pm .000$ \\
$k$-means & $.913 \pm .000$ & $.850 \pm .000$ & $.491 \pm .000$ & $.464 \pm .000$ \\
AC & $\underline{.933} \pm .000$ & $\underline{.878} \pm .000$ & $\boldsymbol{.689} \pm .000$ & $\boldsymbol{.568} \pm .000$ \\
DBSCAN & $.854 \pm .009$ & $.745 \pm .011$ & $.554 \pm .018$ & $.447 \pm .010$ \\
GMM & $.909 \pm .000$ & $.846 \pm .000$ & $\underline{.664} \pm .000$ & $\underline{.558}\pm .000$ \\
\bottomrule
\end{tabular}
\end{center}
\end{table}

Table~\ref{tab:mnist-cancer-cmp} shows the results.  We repeat each experiment five times and report the mean $\pm$ standard deviation.  We highlight each metric's first and second highest values in bold and underlined.  For MNIST, the top 3 methods are our MBMM, followed by AC, and then $k$-means. For the cancer dataset, the best performance is achieved using the AC algorithm, followed by our MBMM and GMM. In general, our MBMM and AC perform best among all.

\subsection{Distance between data points}

\figwidth{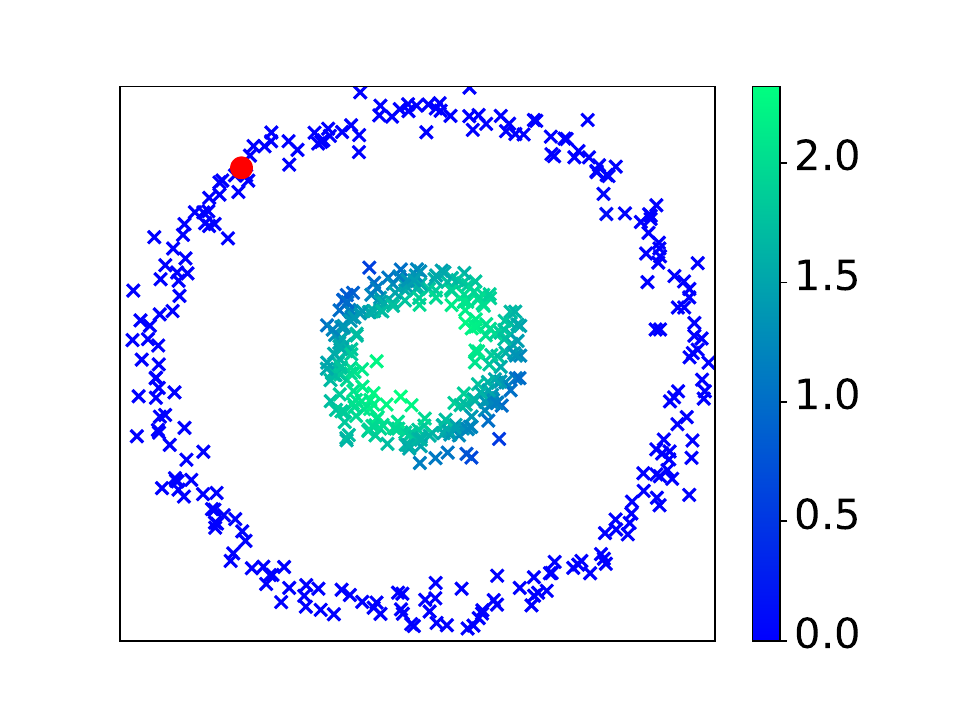}{Distance from red point to other points in concentric circles dataset}{fig:point-dist}{.5 \textwidth}

As explained in Section~\ref{sec:sim-func}, we define the distance between $\boldsymbol{x}_i$ and $\boldsymbol{x}_j$ based on the KL divergence between $[\gamma_{i,1},\ldots, \gamma_{i,C}]^T$ and $[\gamma_{j,1},\ldots, \gamma_{j,C}]^T$. Consequently, even if $\boldsymbol{x}_i$ and $\boldsymbol{x}_j$ are distant based on the Euclidean distance, they could still have a small distance score if $\gamma_{i,c} \approx \gamma_{j,c}$ for most $c$-s.  

We illustrate the red data point's distances to others using concentric circles in Figure~\ref{fig:point-dist}. Outer circle points are closer, showing that MBMM's distance function can assign small values even with large Euclidean distances.


\section{Related Work} \label{sec:rel-work}

\begin{table*}[tb]
\caption{Popular clustering algorithms and their properties}
\label{tab:cluster-algo-cmp}
\begin{center}
\begin{tabular}{c||ccccc}
\toprule
 & \textbf{MBMM} & \textbf{$k$-means} & \textbf{DBSCAN} & \textbf{AC} & \textbf{GMM} \\
\hline \hline
Type & \makecell[c]{Distribution-\\based} & \makecell[c]{Centroid-\\based} & \makecell[c]{Density-\\based} & \makecell[c]{Hierarchical\\clustering} & \makecell[c]{Distribution-\\based} \\
Assignment        & Soft & Hard & Hard & Hard & Soft \\
Cluster shape          & Versatile  & Convex     & Versatile & Versatile & Convex     \\
Generative/Discriminative & Gen. & Discr. & Discr. & Discr. & Gen. \\
\bottomrule
\end{tabular}
\end{center}
\end{table*}

Clustering algorithms can be classified into four types based on how they partition data points: hierarchical, centroid-based, density-based, and distribution-based clustering.

Hierarchical clustering algorithms are top-down or bottom-up, corresponding to the iterative division of each cluster into smaller clusters and the aggregation of smaller clusters into larger clusters. Hierarchical clustering algorithms allow dynamically adjusting the cluster numbers. However, users must define the distance between not only data points but also between clusters, which could sometimes be counterintuitive. Well-known hierarchical clustering algorithms include agglomerative clustering (AC) and BIRCH~\cite{zhang1996birch}.

Centroid-based algorithms represent each cluster using a centroid, assigning each data point to the closest cluster. However, these algorithms generate clusters with convex shapes, eliminating the possibility of fitting a bimodal cluster. Well-known algorithms include $k$-means, $k$-medoids, $k$-medians, and $k$-means$++$.

Density-based algorithms determine clusters by assuming that densely distributed areas are clusters. Typical algorithms include DBSCAN~\cite{ester1996density} and OPTICS~\cite{ankerst1999optics}.  Although they discover clusters of various shapes, hyperparameter tuning can be time-consuming and heavily influence clustering results~\cite{karami2014choosing}. Additionally, density-based algorithms sometimes have difficulty clustering data points when the distances between different data points vary widely.

Distribution-based models assume that each cluster follows a probability distribution. One well-known is GMM, which assumes that each cluster follows a Gaussian distribution. Distribution-based models naturally generate synthetic data points by sampling a cluster ID and then one data point from the data distribution of cluster~$i$. One problem with GMM is that the shape of each cluster must be convex since this is a fundamental property of a Gaussian distribution. As a result, GMM cannot differentiate between inner and outer
circles.

Table~\ref{tab:cluster-algo-cmp} gives an overview of these clustering algorithms and their properties. Both MBMM and GMM are distribution-based models, which thus allow for soft clustering and synthetic generation of the data points. DBSCAN and AC allow non-convex cluster shapes in which each data point within a cluster is close to only a few data points from the same cluster. MBMM also supports non-convex cluster shapes due to the versatility of the multivariate beta distribution. 

The beta mixture model has been studied in the bioinformatics and biochemical domains~\cite{ji2005applications,schroder2017hybrid}. However, they assumed that each cluster follows a standard beta distribution, which limits the practical usage of these models because each data point must be univariate. This constraint is probably due to the fact that the definition of a multivariate beta distribution is still ambiguous. Our study is more practical because we allow each data point to be multivariate.

\section{Discussion} \label{sec:disc}

This paper proposes a new probabilistic model, the multivariate beta mixture model, for data clustering. We demonstrate MBMM's effectiveness by thorough experiments on synthetic and real datasets. Furthermore, MBMM is a generative model that allows for the generation of new data points. Compared to another famous generative clustering algorithm, the Gaussian mixture model, MBMM allows for a more flexible cluster shape. To ensure reproducibility, we have released our experimental code and encapsulated the MBMM module as a class with typical class methods supported by the clustering algorithms in \texttt{scikit-learn}, facilitating the utilization of the MBMM in various applications.

Although MBMM has these nice properties, we believe that different clustering algorithms should be used in combination to jointly partition data points for the following reasons. First, data clustering is ill-defined due to the lack of ground truth labels during both training and testing, making the choice of training objective and evaluation ad hoc~\cite{caruana2006meta}.  Additionally, it has been shown that, under reasonably general conditions, no single clustering algorithm can satisfy the three fundamental properties introduced in~\cite{kleinberg2003impossibility}.

The capacity of MBMM is limited by the need for a positive correlation among all variates due to parameter $b$~\cite{jones2002multivariate}. We plan to explore multivariate beta distributions allowing both positive and negative correlations based on~\cite{arnold2011flexible}.


\bibliographystyle{unsrt}  
\bibliography{ref}

\end{document}